\documentclass[10pt, a4paper]{article}

\usepackage[final]{lrec2026} 

\usepackage{multirow}
\usepackage{arydshln}
\usepackage{xurl}
\usepackage{xspace}
\usepackage{verbatim}

\newcommand{\corpusname}{\texttt{PARCOMED}\xspace}
\newcommand{\githublink}{\url{https://github.com/PARTAGES-dev/partages-llm}\xspace}
\newcommand{\hdhlink}{\url{https://huggingface.co/HealthDataHub/PARCOMED}\xspace}

\title{Is Biomedical Specialization Still Worth It? Insights from Domain-Adaptive Language Modelling with a New French Health Corpus}

\name{\bfseries
Aidan Mannion$^{1}$,
Cécile Macaire$^{1}$,
Armand Violle$^{2}$,
Stéphane Ohayon$^{2}$,\\\large \bfseries
Xavier Tannier$^{2}$,
Didier Schwab$^{1}$,
Lorraine Goeuriot$^{1}$,
François Portet$^{1}$
}

\address{
$^{1}$ Université Grenoble Alpes, CNRS, Grenoble INP, LIG, 38000 Grenoble, France \\
$^{2}$ Sorbonne Université, LIMICS, 15 rue de l'École de Médecine, 75006 Paris, France
}

\abstract{
    Large language models (LLMs) have demonstrated remarkable capabilities across diverse domains, yet their adaptation to specialized fields remains challenging, particularly for non-English languages.
    This study investigates domain-adaptive pre-training (DAPT) as a strategy for specializing small to mid-sized LLMs in the French biomedical domain through continued pre-training.
    We address two key research questions: the viability of specialized continued pre-training for domain adaptation and the relationship between domain-specific performance gains and general capability degradation.
    Our contributions include the release of a fully open-licensed French biomedical corpus suitable for commercial and open-source applications, the training and release of specialized French biomedical LLMs, and novel insights for DAPT implementation. 
    Our methodology encompasses the collection and refinement of high-quality French biomedical texts, the exploration of causal language modeling approaches using DAPT, and conducting extensive comparative evaluations.
    Our results cast doubt on the efficacy of DAPT, in contrast to previous works, but we highlight its viability in smaller-scale, resource-constrained scenarios under the right conditions.
    Our findings further suggest that model merging post-DAPT is essential to mitigate generalization trade-offs, and in some cases even improves performance on specialized tasks at which the DAPT was directed.
\\ \newline 
}

\begin{document}

\maketitleabstract

\section{Introduction}
\label{sec:intro}

LLMs are widely recognized as \textit{foundation models} that demonstrate promising general capabilities, often exhibiting emergent reasoning abilities with appropriate prompting \citep{bommasani_opportunities_2021}.
However, achieving high performance and clinical reliability in specialized areas requires thoughtful adaptation.
Domain-Adaptive Pre-training (DAPT, \citealp{gururangan_dont_2020}), also referred to as Continual Pre-Training (CPT, \citealp{chen_towards_2025}), addresses this by conducting a second phase of pre-training on large, unlabeled, domain-specific text to align the model with the distributional characteristics of text in the target field.
This approach aims to capture useful patterns, such as complex medical terminology, that may be inadequately represented in the initial, broad general-purpose training corpus.

The Domain-Adaptive Pre-Training presented in this work is carried out as part of the of the R\&D phase of the PARTAGES project, which aims to develop specialized language models for use in the automation of document-processing tasks in the French healthcare system, while releasing the associated resources (models, code, datasets) as freely-available open-source tools.

In this context, we present a new collection of French biomedical corpora that is guaranteed to be fully compatible with all downstream applications from a licensing standpoint, called \corpusname (\textbf{PAR}TAGES \textbf{C}orpus of \textbf{O}pen \textbf{ME}dical \textbf{D}ocuments).
Alongside the corpus, we release a collection of domain-specialized models trained thereon, using Qwen3 \citep{yang_qwen3_2025} as a foundation, and reflect on the utility of this kind of continual pre-training as an efficacious strategy going forward.

\section{Related Work}
\label{sec:related-work}
The application of LLMs to medicine has resulted in several high-profile models, predominantly in English, trained via proprietary or open-source DAPT methodologies, often relying on massive datasets of biomedical literature.
Google's Med-PaLM \citep{singhal_large_2023}, for example, built upon a 540-billion parameter foundation model, achieved state-of-the-art results on medical question-answering benchmarks by combining scaling with prompt tuning strategies.
Open-source alternatives have also emerged, focusing on scalability and accessibility, such as BioMedLM (\citealp{bolton_biomedlm_2024}; 2.7B parameters), BioGPT (\citealp{luo_biogpt_2022}; 355M), and MedAlpaca (\citealp{han_medalpaca_2023}; 7B \& 13B).
Another significant open-source contribution is MEDITRON \citep{chen_meditron-70b_2023}, which scaled medical CPT to 70B parameters using Llama-2 as a backbone, training on a corpus that included PubMed abstracts, full-text papers, and high-quality clinical guidelines.
Similarly, BioMistral-7B \citep{labrak_biomistral_2024} leveraged the Mistral-7B-Instruct model, supplementing its training with the PubMed Central Open Access Subset to specialize it for the biomedical domain.
These foundational English models, along with related encoder-only models like BioBERT \citep{lee_biobert_2020}, established that CPT has the potential to enhance medical-specific language modelling capabilities in certain scenarios.

Despite the reported gains, the necessity of DAPT for highly capable, general-purpose LLMs has been challenged.
Recent head-to-head comparisons, using rigorous evaluation protocols that involve optimizing prompts for each model independently and measuring statistical significance, found that most biomedical LLMs failed to consistently improve over their general-domain base models in zero- or few-shot QA tasks \citep{jeong_medical_2024}.

For domains outside of English, such as the French biomedical context, the challenges are magnified by the scarcity of specialized resources.
Multilingual generalization remains limited, as performance typically degrades when models are tested on automatically translated benchmarks, as shown by \citet{labrak_biomistral_2024}, who also highlighted that additional pre-training on English medical data has limited benefits for non-English contexts.
Addressing the French medical domain specifically, researchers have introduced specialized resources for CPT like the NACHOS corpus \citep{labrak_drbert_2023} and the automatically-translated TransCorpus-bio-fr \citep{knafou_transbert_2025}, recognizing that data scarcity is a major hurdle in releasing open-source specialized LLMs in French.
A promising direction for more comprehensive evaluations of these strategies is the systematic testing of CPT, SFT (Supervised Fine-Tuning), and combined CPT+SFT approaches, such as the work by \citet{belmadani_adaptation_2025} on the Mistral-7B architecture.

\section{The \corpusname Corpus}
\label{sec:corpora}
\subsection{Context}

The availability of French biomedical data remains a major challenge for improving the multilingual capabilities of large language models (LLMs) in the medical domain.

We introduce and release the \corpusname corpus, a comprehensive collection of French biomedical texts compiled from a wide range of sources.
Although collections of French medical documents, such as NACHOS~\citep{labrak_drbert_2023} or Jargon~\citep{segonne:hal-04535557} have already been distributed to the community recently, our corpus collection is the result of a greater scrutiny of the licensing term of each source.
Thus, in contrast to the collections mentioned above, the \corpusname corpus is fully compatible with research usage and is also distributed with a version compatible with commercial usage.

The selected datasets for our corpus come from a variety of sources which can be categorized as follows (for readability, citations are provided in Table~\ref{tab:corpus}):
\begin{itemize}
    \item Open-access archives (HAL, HAS, ISTEX, ANSES, QUALISCOPE, CERIMES, CNEDIMTS, ECDC TM).
    \item Healthcare data such as clinical cases from the E3C, CAS (real, anonymized cases), and FRASIMED (synthetic) corpora, as well as clinical trial protocols (ESSAI).
    \item Information leaflets for medications (BDPM, EMEA V3).
    \item Datasets available in literature designed for specific NLP tasks such as machine translation (WMT16, WMT18 Medline), named-entity recognition (QUAERO, DEFT2021, CLEAR, MANTRA GSC), multiple-choice QA (FrenchMedMCQA) and doctor-patient dialogues (MQC, PXCORPUS).
    \item General knowledge on health and medicine extracted via API requests\footnote{\url{https://wikipedia-api.readthedocs.io/en/latest/}} to French Wikipedia for medecine, pharmacy and biology categories.
\end{itemize}

\subsection{Data collection}

As mentioned previously, our sources cover diverse biomedical content, including scientific articles, drug leaflets, medical device evaluations, regulatory documents, clinical case reports, and institutional recommendations. In each case, all partitions (train/dev/test) of the datasets were included. We provide two distinct versions of the aggregated dataset, summarized in Table~\ref{tab:corpus}: a commercial-use corpus, containing only sources whose licenses permit commercial use, and a research-only corpus, allowing only non-commercial applications. As it can be seen, the corpus is dominated by scientific documents (around 94\% of words).

\begin{table*}[ht]
    \small
    \centering
    \renewcommand{\arraystretch}{1.2}
    \begin{tabular}{lccrrp{10em}}
        \hline
        \textbf{Source name} & \textbf{Document type} & \textbf{Commercial} & \textbf{\# docs} & \textbf{\# words} & \textbf{Reference} \\
        \hline
        HAL             & Scientific & Yes   & 26,987    & 703,473,770   & \citet{hal} \\
        HAS             & Scientific & Yes   & 11,334    & 96,173,390    & \citet{has} \\
        ISTEX           & Scientific & Yes   & 12,179    & 43,138,368    & \citet{istex} \\
        BDPM            & Medication & Yes   & 11,026    & 20,035,903    & \citet{bdpm} \\
        WIKIPEDIA       & Encyclopedic          & Yes   & 9,957     & 6,531,021     & \citet{wikipedia} \\
        WMT16           & Scientific & Yes   & 587,075   & 6,490,287     & \citet{bojar2016findings}         \\
        EMEA V3         & Medication & Yes   & 222,971   & 4,449,136     & \citet{tiedemann-2012-parallel}   \\
        CERIMES         & Educational & Yes   & 22        & 1,715,189     & \citet{cerimes}                   \\
        FRASIMED        & Clinical & Yes   & 2,048     & 1,322,895     & \citet{zaghir2024frasimed}        \\
        DEFT2021        & Question Answering & Yes   & 271       & 110,641       & \citet{grouin2021classification}  \\
        QUAERO          & Scientific & Yes   & 2,490     & 71,812        & \citet{neveol2014quaero}          \\ 
        FrenchMedMCQA   & Question Answering & Yes   & 1,144     & 58,872        & \citet{labrak2022frenchmedmcqa}   \\
        CNEDIMTS        & Regulation & Yes   & 813       & 58,345        & \citet{cnedimts}                  \\
        ECDC TM         & Other medical & Yes   & 2,160     & 42,491        & \citet{steinberger2014overview}   \\
        PXCORPUS        & Medication & Yes   & 1,414     & 18,372        & \citet{kocabiyikoglu2022spoken}   \\
        QUALISCOPE      & Regulation & Yes   & 298       & 11,736        & \citet{qualiscope}                \\
        MANTRA GSC      & Scientific & Yes   & 150       & 3,596         & \citet{kors2015multilingual}      \\
        \hline
        \textbf{Total commercial}   & & & \textbf{892,343} & \textbf{883,706,984}   \\
        \hline
        E3C             & Clinical & No    & 7,499     & 15,864,637    & \citet{Minard2021European}        \\
        CAS             & Clinical & No    & 716       & 233,371       & \citet{grabar2018cas}             \\
        CLEAR           & Scientific & No    & 6         & 226,123       & \citet{grabar:halshs-01968355}    \\
        ESSAI           & Clinical & No    & 5,842     & 146,537       & \citet{Dalloux2021}               \\
        MQC             & Dialogue & No    & 38        & 15,672        & \citet{laleye2020french}          \\
        WMT18 Medline   & Scientific & No    & 49        & 7,719         & \citet{neves-etal-2018-findings}  \\
        \hline
        \textbf{Total research} & & & \textbf{906,489} & \textbf{900,199,883} \\
        \hline
    \end{tabular}
    \caption{Data sources for the \corpusname corpus.}
    \label{tab:corpus}
\end{table*}


\begin{table*}[ht]
    \small
    \centering
    \renewcommand{\arraystretch}{1.1}
    \begin{tabularx}{\textwidth}{r|l|l|c|c|c}
    \textbf{Task Group}              & \textbf{Topic} & \textbf{Abbreviation} & \textbf{Source} & \textbf{Eval. Metric} & \textbf{\# Questions} \\
    \hline
    \multirow{7}{*}{(EN/FR)-MEDICAL} & Anatomy & Anat.                 & MMLU & Accuracy & 135 \\
                                     & Clinical Knowledge & C.K.       & MMLU & Accuracy & 265 \\
                                     & College Biology & CBio.         & MMLU & Accuracy & 144 \\
                                     & College Medicine & CMed.        & MMLU & Accuracy & 173 \\
                                     & Health  & n/a                   & MMLU-Pro-X & Exact match & 687 \\
                                     & Medical Genetics & MGen.        & MMLU & Accuracy & 100 \\
                                     & Professional Medicine & ProMed. & MMLU & Accuracy & 272 \\
    \hline
    \multirow{7}{*}{(EN/FR)-OTHER}   & Business & Bus.                 & MMLU-Pro-X & Exact match & 789 \\
                                     & Computer Science & CS           & MMLU-Pro-X & Exact match & 410 \\
                                     & Economics & Econ.               & MMLU-Pro-X & Exact match & 844 \\
                                     & History & Hist.                 & MMLU-Pro-X & Exact match & 381 \\
                                     & Law      & n/a                  & MMLU-Pro-X & Exact match & 959 \\
                                     & Philosophy & Phil.              & MMLU-Pro-X & Exact match & 499 \\
                                     & Psychology & Psych.             & MMLU-Pro-X & Exact match & 798 \\
    \hline
    \end{tabularx}
    \caption{Groupings, abbreviations, metrics and number of questions for each the QA datasets used for evaluation.}
    \label{tab:eval-datasets}
\end{table*}

\subsection{Text cleaning and volume}

All documents were preprocessed using a the pipeline inspired by~\citet{le2020flaubert}, including Unicode conversion and normalization, removal of characters outside standard French encoding, suppression of multiple spaces, and deletion of URLs.
The dataset is organized at the document level, where each entry corresponds to a single document (e.g., a Wikipedia page). In total, 906,489 documents were collected from various sources (see Table~\ref{tab:corpus}); the corpus used to train the models was the 892K-document version allowing commercial use.

\section{Domain-Adaptive Continual Pre-Training for Medical Applications in French}
\label{sec:training}

The experimental methodology discussed in this paper proceeds in three main steps: model selection, DAPT, and merging.
Firstly, we run a range of baseline evaluations and selected the best-performing generalist foundation models for DAPT (the evaluation protocol is presented in Section \ref{sec:eval}).
We then run Causal Language Modelling on these models, executing the evaluation benchmark at regular intervals.
Based on the progression of the averaged evaluation metrics, we select a checkpoint to focus on in the final results.
Finally, using Spherical Linear Interpolation (SLERP,~\citet{goddard_arcees_2024}), we combine the weights of this checkpoint with the base model, in order to investigate the resulting trade-offs in evaluation results.
Evaluation results for the selected checkpoint and its corresponding SLERP merge are presented in Section \ref{sec:results}.

\paragraph{Model Selection} The generalist foundation models used in these experiments are from the Qwen3 family. 

Having implemented the evaluation of a range of decoder language models on a broad bilingual multi-domain question-answering benchmark, of which a subset is presented in Section \ref{sec:results}, the 8B model stood out as the best-performing base LLM, surpassing not only direct competitors from the Llama and Mistral families, but also domain-specific models such as Apollo-7B \citep{zheng_efficiently_2024}, and BioMistral \citep{labrak_biomistral_2024}.
``Best-performing'' in this context refers to the model ranking on a selection of biomedical tasks in French (our target domain), for which more complete results can be found in Appendix \ref{sec:apdx-benchmark}.
To investigate the effect of model size on DAPT in this context, we also carry out all of our experiments using three other Qwen3 models, the 0.6B, 1.7B, and 4B variants.

We restrict our attention in this work to the ``-Base'' variants, which have not undergone instruction tuning.
This choice was made in order to more reliably isolate the effects of unsupervised training.
In addition, we aim to further fine-tune our domain-specialized models on medical  document-processing use cases for which the conversational ``chatbot-like'' behaviour inculcated by instruction tuning is not necessarily desirable.

\subsection{Continual Pretraining Setup: PDAPT}

After tokenizing the \corpusname commercial corpus and chunking it into sequences of 2,048 tokens (longer documents were split with an overlap stride of 4 tokens), we continue the pre-training of the four Qwen3 base models for a total of 4,320 update steps.
The tokenized corpus contains over 1.95B tokens\footnote{1,955,165,272} from a word count of uner 1B, pointing to the large amount of specialized domain-specific terminology contained therein.

This training was carried out with a constant learning rate of $2\times10^{-5}$ with no warmup, and an effective batch size (taking into account gradient accumulation and data parallelism) of 1,152 sequences.
The full training run thus corresponds to 2.53 epochs over the corpus.
The progressive effect of DAPT on downstream task performance is investigated by checkpointing the training state every 720 steps (see Figure \ref{fig:linechart}).
Training runs were executed on 48 NVIDIA H100 GPUs on the Jean Zay computing cluster, using BF16 precision and the Fully Sharded Data Parallel framework from PyTorch.

We abbreviate this continual pretraining process as PDAPT (\corpusname\texttt{DAPT}).

\begin{table*}[ht]
    \centering
    \setlength{\tabcolsep}{5pt}
    \renewcommand{\arraystretch}{1.1}
    \begin{tabularx}{\textwidth}{l|ccccccc}
        \textbf{Subject} $\rightarrow$ & \textbf{Anat.} & \textbf{C.K.} & \textbf{CBio.} & \textbf{CMed.} & \textbf{Health} & \textbf{MGen.} & \textbf{ProMed.} \\
        \hline
        Qwen3-0.6B-Base & 30.4$\pm$4.0 & 49.8$\pm$3.1 & 39.6$\pm$4.1 & 42.8$\pm$3.8 & 15.9$\pm$1.4 & 48.0$\pm$5.0 & 37.1$\pm$2.9 \\
        +PDAPT & \cellcolor{green!25}\textbf{39.3$\pm$4.2} & \textbf{50.2$\pm$3.1} & \textbf{41.0$\pm$4.1} & \cellcolor{green!25}\textbf{49.1$\pm$3.8} & \textbf{16.9$\pm$1.4} & \textbf{48.0$\pm$5.0} & \cellcolor{green!25}\textbf{44.5$\pm$3.0} \\
        +SLERP & \cellcolor{green!25}\textbf{43.7$\pm$4.3} & 49.8$\pm$3.1 & \cellcolor{green!25}\textbf{43.8$\pm$4.1} & 42.8$\pm$3.8 & \cellcolor{green!25}\textbf{18.0$\pm$1.5} & 44.0$\pm$5.0 & \textbf{37.9$\pm$2.9} \\
        \cdashline{1-8}
        Qwen3-1.7B-Base & 48.1$\pm$4.3 & 59.2$\pm$3.0 & 54.2$\pm$4.2 & 59.5$\pm$3.7 & 25.9$\pm$1.7 & 66.0$\pm$4.8 & 56.2$\pm$3.0 \\
        +PDAPT & \cellcolor{green!25}\textbf{58.5$\pm$4.3} & \textbf{59.2$\pm$3.0} &\textbf{60.4$\pm$4.1} & 57.8$\pm$3.8 & 24.3$\pm$1.6 & 62.0$\pm$4.9 & \textbf{57.7$\pm$3.0} \\
        +SLERP & \textbf{51.9$\pm$4.3} &\textbf{ 61.1$\pm$3.0} & \textbf{60.4$\pm$4.1} & \textbf{61.8$\pm$3.7} & \textbf{26.2$\pm$1.7} & \textbf{67.0$\pm$4.7} & \textbf{59.6$\pm$3.0} \\
        \cdashline{1-8}
        Qwen3-4B-Base & 54.8$\pm$4.3 & 70.9$\pm$2.8 & 75.0$\pm$3.6 & 68.2$\pm$3.6 & 39.3$\pm$1.9 & 74.0$\pm$4.4 & 69.5$\pm$2.8 \\
        +PDAPT & \textbf{58.5$\pm$4.3} & 70.6$\pm$2.8 & \textbf{77.8$\pm$3.5} & \textbf{71.1$\pm$3.5} & 37.0$\pm$1.8 & \textbf{81.0$\pm$3.9} & \textbf{71.7$\pm$2.7} \\
        +SLERP & \textbf{57.0$\pm$4.3} & \textbf{74.0$\pm$2.7} & \textbf{78.5$\pm$3.4} & \textbf{71.1$\pm$3.5} & \textbf{40.6$\pm$1.9} & \textbf{79.0$\pm$4.1} & \textbf{72.4$\pm$2.7} \\
        \cdashline{1-8}
        Qwen3-8B-Base & 62.2$\pm$4.2 & 74.7$\pm$2.7 & 87.5$\pm$2.8 & 75.7$\pm$3.3 & 50.2$\pm$1.9 & 80.0$\pm$4.0 & 76.5$\pm$2.6 \\
        +PDAPT & 61.5$\pm$4.2 & \textbf{76.2$\pm$2.6} & 86.8$\pm$2.8 & \textbf{76.9$\pm$3.2} & \cellcolor{red!25}45.9$\pm$1.9 & 80.0$\pm$4.0 & 76.1$\pm$2.6 \\
        +SLERP & 60.0$\pm$4.2 & \textbf{77.4$\pm$2.6} & 86.8$\pm$2.8 & \textbf{76.3$\pm$3.2} & 49.8$\pm$1.9 & 79.0$\pm$4.1 & 75.7$\pm$2.6 \\
        \hline
        \end{tabularx}
    \caption{Comparative accuracy scores for the task group FR-MEDICAL.}
    \label{tab:fr-medical}
\end{table*}

\begin{table*}[ht]
    \centering
    \setlength{\tabcolsep}{5pt}
    \renewcommand{\arraystretch}{1.1}
    \begin{tabularx}{\textwidth}{l|ccccccc}
        \textbf{Subject} $\rightarrow$ & \textbf{Anat.} & \textbf{C.K.} & \textbf{CBio.} & \textbf{CMed.} & \textbf{Health} & \textbf{MGen.} & \textbf{ProMed.} \\
        \hline
        Qwen3-0.6B-Base & 47.4$\pm$4.3 & 57.0$\pm$3.0 & 59.7$\pm$4.1 & 52.6$\pm$3.8 & 22.4$\pm$1.6 & 62.0$\pm$4.9 & 55.5$\pm$3.0 \\
        +PDAPT & 40.7$\pm$4.2 & 51.3$\pm$3.1 & 59.0$\pm$4.1 & 51.4$\pm$3.8 & \cellcolor{red!25}18.0$\pm$1.5 & \cellcolor{red!25}52.0$\pm$5.0 & 50.7$\pm$3.0 \\
        +SLERP & 41.5$\pm$4.3 & 54.0$\pm$3.1 & 59.7$\pm$4.1 & \textbf{53.2$\pm$3.8} & 22.0$\pm$1.6 & 59.0$\pm$4.9 & 53.7$\pm$3.0 \\
        \cdashline{1-8}
        Qwen3-1.7B-Base & 59.3$\pm$4.2 & 67.9$\pm$2.9 & 72.9$\pm$3.7 & 68.2$\pm$3.6 & 34.6$\pm$1.8 & 73.0$\pm$4.5 & 64.7$\pm$2.9 \\
        +PDAPT & 57.8$\pm$4.3 & \textbf{68.7$\pm$2.9} & \textbf{74.3$\pm$3.7} & 65.9$\pm$3.6 & \cellcolor{red!25}30.3$\pm$1.8 & 69.0$\pm$4.6 & 59.9$\pm$3.0 \\
        +SLERP & 59.3$\pm$4.2 & 67.9$\pm$2.9 & \textbf{73.6$\pm$3.7} & 67.1$\pm$3.6 & \textbf{35.7$\pm$1.8} & 71.0$\pm$4.6 & 63.6$\pm$2.9 \\
        \cdashline{1-8}
        Qwen3-4B-Base & 68.1$\pm$4.0 & 80.4$\pm$2.4 & 84.7$\pm$3.0 & 74.0$\pm$3.3 & 49.9$\pm$1.9 & 81.0$\pm$3.9 & 78.3$\pm$2.5 \\
        +PDAPT & 64.4$\pm$4.1 & \textbf{80.8$\pm$2.4} & \textbf{86.1$\pm$2.9} & \textbf{74.6$\pm$3.3} & 46.3$\pm$1.9 & \textbf{83.0$\pm$3.8} & 75.7$\pm$2.6 \\
        +SLERP & \textbf{69.6$\pm$4.0} & 80.4$\pm$2.4 & \textbf{86.1$\pm$2.9} & \textbf{75.7$\pm$3.3} & 48.9$\pm$1.9 & \textbf{83.0$\pm$3.8} & \textbf{79.4$\pm$2.5} \\
        \cdashline{1-8}
        Qwen3-8B-Base & 74.1$\pm$3.8 & 80.0$\pm$2.5 & 88.9$\pm$2.6 & 78.0$\pm$3.2 & 55.8$\pm$1.9 & 86.0$\pm$3.5 & 83.5$\pm$2.3 \\
        +PDAPT & 70.4$\pm$3.9 & 78.9$\pm$2.5 & 84.7$\pm$3.0 & 76.9$\pm$3.2 & 55.0$\pm$1.9 & 85.0$\pm$3.6 & 80.9$\pm$2.4 \\
        +SLERP & 72.6$\pm$3.9 & \textbf{81.5$\pm$2.4} & 88.9$\pm$2.6 & 76.9$\pm$3.2 & \textbf{57.1$\pm$1.9} & 86.0$\pm$3.5 & 81.6$\pm$2.4 \\
        \hline
    \end{tabularx}
    \caption{Comparative results for the task group EN-MEDICAL.}
    \label{tab:en-medical}
\end{table*}

\begin{table*}[ht]
    \centering
    \setlength{\tabcolsep}{5pt}
    \renewcommand{\arraystretch}{1.1}
    \begin{tabularx}{\textwidth}{l|ccccccc}
        \textbf{Subject} $\rightarrow$ & \textbf{Bus.} & \textbf{CS} & \textbf{Econ.} & \textbf{Hist.} & \textbf{Law} & \textbf{Phil.} & \textbf{Psych.} \\
        \hline
        Qwen3-0.6B-Base & 19.1$\pm$1.4 & 19.5$\pm$2.0 & 23.5$\pm$1.5 & 13.4$\pm$1.7 & 7.3$\pm$0.8 & 17.0$\pm$1.7 & 26.2$\pm$1.6 \\
        +PDAPT & \cellcolor{red!25}15.8$\pm$1.3 & \cellcolor{red!25}8.8$\pm$1.4 & \cellcolor{red!25}16.1$\pm$1.3 & \textbf{14.2$\pm$1.8} & \textbf{8.4$\pm$0.9} & 16.2$\pm$1.7 & \cellcolor{red!25}18.4$\pm$1.4 \\
        +SLERP & \textbf{19.4$\pm$1.4} & \cellcolor{red!25}12.7$\pm$1.6 & 21.4$\pm$1.4 & \textbf{15.0$\pm$1.8} & 7.2$\pm$0.8 & 15.8$\pm$1.6 & \textbf{26.7$\pm$1.6} \\
        \cdashline{1-8}
        Qwen3-1.7B-Base & 35.6$\pm$1.7 & 27.1$\pm$2.2 & 37.2$\pm$1.7 & 18.9$\pm$2.0 & 8.2$\pm$0.9 & 23.0$\pm$1.9 & 38.8$\pm$1.7 \\
        +PDAPT & \cellcolor{red!25}26.2$\pm$1.6 & \cellcolor{red!25}18.8$\pm$1.9 & 33.3$\pm$1.6 & 17.1$\pm$1.9 & 7.5$\pm$0.9 & 20.0$\pm$1.8 & \cellcolor{red!25}32.5$\pm$1.7 \\
        +SLERP & 33.0$\pm$1.7 & \textbf{27.6$\pm$2.2} & 35.5$\pm$1.6 & 17.6$\pm$2.0 & \textbf{9.1$\pm$0.9} & 22.2$\pm$1.9 & 37.1$\pm$1.7 \\
        \cdashline{1-8}
        Qwen3-4B-Base & 50.8$\pm$1.8 & 46.1$\pm$2.5 & 56.8$\pm$1.7 & 34.1$\pm$2.4 & 18.6$\pm$1.3 & 32.1$\pm$2.1 & 54.9$\pm$1.8 \\
        +PDAPT & \cellcolor{red!25}44.4$\pm$1.8 & \cellcolor{red!25}39.8$\pm$2.4 & \cellcolor{red!25}53.1$\pm$1.7 & 30.4$\pm$2.4 & \cellcolor{red!25}15.6$\pm$1.2 & \textbf{35.1$\pm$2.1} & 51.6$\pm$1.8 \\
        +SLERP & \textbf{52.7$\pm$1.8} & 44.9$\pm$2.5 & 56.0$\pm$1.7 & \textbf{35.2$\pm$2.4} & 18.1$\pm$1.2 & \textbf{33.5$\pm$2.1} & 54.5$\pm$1.8 \\
        \cdashline{1-8}
        Qwen3-8B-Base & 61.5$\pm$1.7 & 50.5$\pm$2.5 & 62.7$\pm$1.7 & 40.2$\pm$2.5 & 23.5$\pm$1.4 & 4.9$\pm$2.2 & 60.5$\pm$1.7 \\
        +PDAPT & \cellcolor{red!25}55.4$\pm$1.8 & 45.6$\pm$2.5 & 61.1$\pm$1.7 & 39.4$\pm$2.5 & 21.1$\pm$1.3 & 41.7$\pm$2.2 & 59.3$\pm$1.7 \\
        +SLERP & 58.2$\pm$1.8 & \textbf{53.7$\pm$2.5} & \textbf{63.6$\pm$1.7} & \textbf{41.5$\pm$2.5} & \textbf{23.8$\pm$1.4} & \textbf{45.5$\pm$2.2} & \textbf{61.2$\pm$1.7} \\
        \hline
    \end{tabularx}
	\caption{Comparative exact-match scores for the task group FR-OTHER.}
	\label{tab:fr-other}
\end{table*}

\begin{table*}[ht]
    \centering
    \setlength{\tabcolsep}{5pt}
    \begin{tabularx}{\textwidth}{l|ccccccc}
        \textbf{Subject} $\rightarrow$ & \textbf{Bus.} & \textbf{CS} & \textbf{Econ.} & \textbf{Hist.} & \textbf{Law} & \textbf{Phil.} & \textbf{Psych.} \\
        \hline
        Qwen3-0.6B-Base & 28.8$\pm$1.6 & 26.1$\pm$2.2 & 31.5$\pm$1.6 & 16.0$\pm$1.9 & 11.2$\pm$1.0 & 19.4$\pm$1.8 & 36.6$\pm$1.7 \\
        +PDAPT & \cellcolor{red!25}18.1$\pm$1.4 & \cellcolor{red!25}20.5$\pm$2.0 & \cellcolor{red!25}25.7$\pm$1.5 & 15.5$\pm$1.9 & 10.3$\pm$1.0 & \cellcolor{red!25}15.8$\pm$1.6 & \cellcolor{red!25}27.7$\pm$1.6 \\
        +SLERP & 27.0$\pm$1.6 & 24.9$\pm$2.1 & 30.7$\pm$1.6 & \textbf{17.1$\pm$1.9} & 10.4$\pm$1.0 & 19.0$\pm$1.8 & 34.5$\pm$1.7 \\
        \cdashline{1-8}
        Qwen3-1.7B-Base & 37.1$\pm$1.7 & 39.0$\pm$2.4 & 46.0$\pm$1.7 & 27.6$\pm$2.3 & 14.6$\pm$1.1 & 34.1$\pm$2.1 & 47.0$\pm$1.8 \\
        +PDAPT & \cellcolor{red!25}33.2$\pm$1.7 & 34.4$\pm$2.3 & 43.8$\pm$1.7 & 25.5$\pm$2.2 & \textbf{14.8$\pm$1.1} & \cellcolor{red!25}29.1$\pm$2.0 & \cellcolor{red!25}42.5$\pm$1.8 \\
        +SLERP & \cellcolor{green!25}42.6$\pm$1.8 & 39.0$\pm$2.4 & 45.4$\pm$1.7 & 25.5$\pm$2.2 & 14.2$\pm$1.1 & 32.7$\pm$2.1 & 45.9$\pm$1.8 \\
        \cdashline{1-8}
        Qwen3-4B-Base & 57.3$\pm$1.8 & 53.2$\pm$2.5 & 63.5$\pm$1.7 & 38.6$\pm$2.5 & 25.2$\pm$1.4 & 42.1$\pm$2.2 & 61.8$\pm$1.7 \\
        +PDAPT & \cellcolor{red!25}52.1$\pm$1.8 & \cellcolor{red!25}48.0$\pm$2.5 & 62.1$\pm$1.7 & \textbf{39.6$\pm$2.5} & \cellcolor{red!25}19.2$\pm$1.3 & 40.9$\pm$2.2 & 59.1$\pm$1.7 \\
        +SLERP & 56.3$\pm$1.8 & 51.0$\pm$2.5 & \textbf{65.0$\pm$1.6} & \textbf{39.4$\pm$2.5} & 22.7$\pm$1.4 & \textbf{43.3$\pm$2.2} & 61.3$\pm$1.7 \\
        \cdashline{1-8}
        Qwen3-8B-Base & 62.5$\pm$1.7 & 60.2$\pm$2.4 & 68.2$\pm$1.6 & 48.3$\pm$2.6 & 29.6$\pm$1.5 & 49.3$\pm$2.2 & 67.2$\pm$1.7 \\
        +PDAPT & 60.7$\pm$1.7 & 59.8$\pm$2.4 & 68.2$\pm$1.6 & \textbf{49.9$\pm$2.6} & 27.7$\pm$1.4 & 47.1$\pm$2.2 & 66.0$\pm$1.7 \\
        +SLERP & \textbf{62.6$\pm$1.7} & 58.3$\pm$2.4 & 67.4$\pm$1.6 & \textbf{49.9$\pm$2.6} & \textbf{29.7$\pm$1.5} & \textbf{50.7$\pm$2.2} & 66.2$\pm$1.7 \\
        \hline
    \end{tabularx}
	\caption{Comparative results for the task group EN-OTHER.}
	\label{tab:en-other}
\end{table*}

\section{Evaluation Protocol}
\label{sec:eval}

The evaluation methodology presented in this paper relies on a set of standardized LLM evaluation benchmarks in both English and French.
The specific aims of this evaluation framework are firstly to evaluate whether or not specializing LLMs from the general domain improves their performance on biomedical tasks, and secondly to compare PDAPT model performance on general-purpose benchmarks with their corresponding base models to identify potential degradation due to over-specialization.

The evaluation is based around the open-source framework ``lm-evaluation-harness''\footnote{\url{https://github.com/EleutherAI/lm-evaluation-harness}} \citep{eval-harness} for few-shot language model assessment, which ensures full reproducibility through open and publicly available datasets.
In order to measure the trade-off between specialization and generalization brought about by the DAPT strategy outlined in Section \ref{sec:training}, we define four task groups: one in the target domain (medicine) and the target language (French), one in the target domain in a different language (English) and two more that constitute a collection of other specialized domains outside of medicine, in both languages.
Each group contains seven tasks, laid out in Table \ref{tab:eval-datasets}.
The evaluation datasets themselves are drawn from two sources:
\begin{itemize}
    \item The MMLU multiple-choice question-answering dataset \citep{hendrycks_measuring_2021}, from which we draw a selection of medical-domain tasks; for French-language evaluation, we reuse the translated versions from \citet{labrak_biomistral_2024}.
    \item The MMLU-Pro-X dataset \citep{xuan_mmluprox_2025}, a diverse multilingual benchmark built to evaluate the reasoning capacities of LLMs.
\end{itemize}

We reuse the standard task configuration and metrics for these tasks, as integrated in lm-evaluation-harness: for the medical MMLU tasks, we use few-shot prompting with $n=3$ and use accuracy as the evaluation metric, while for MMLU-Pro-X, we use $n=5$ and the \textit{exact-match} metric.
The first of these metrics, referred to as ``Accuracy'' in Table \ref{tab:eval-datasets}, considers a model's answer to be the the string with the highest conditional log probability from a fixed set of possible answer strings.
The exact-match metric, on the other hand, only considers the overall highest-probability string to be the answer.
In both cases, the aggregate metric corresponds to the percentage of model answers that match the ground-truth label.
Each of these metrics is accompanied by a confidence interval based on a bootstrapped standard error measurement implemented via the evaluation harness; as can be seen in Section \ref{sec:results}'s tables, the smaller dataset sizes for the medical-specific tasks result in wider intervals in general.

As a summary statistic for the general performance tendencies at the level of our four task groups, we calculate an average of these metrics weighted by the number of documents in each dataset.
This metric is referred to simply as the ``weighted average score'' in Section \ref{sec:results}.

\section{Results and Analysis}
\label{sec:results}

\begin{figure}[ht]
    \begin{center}
        \includegraphics[width=\columnwidth]{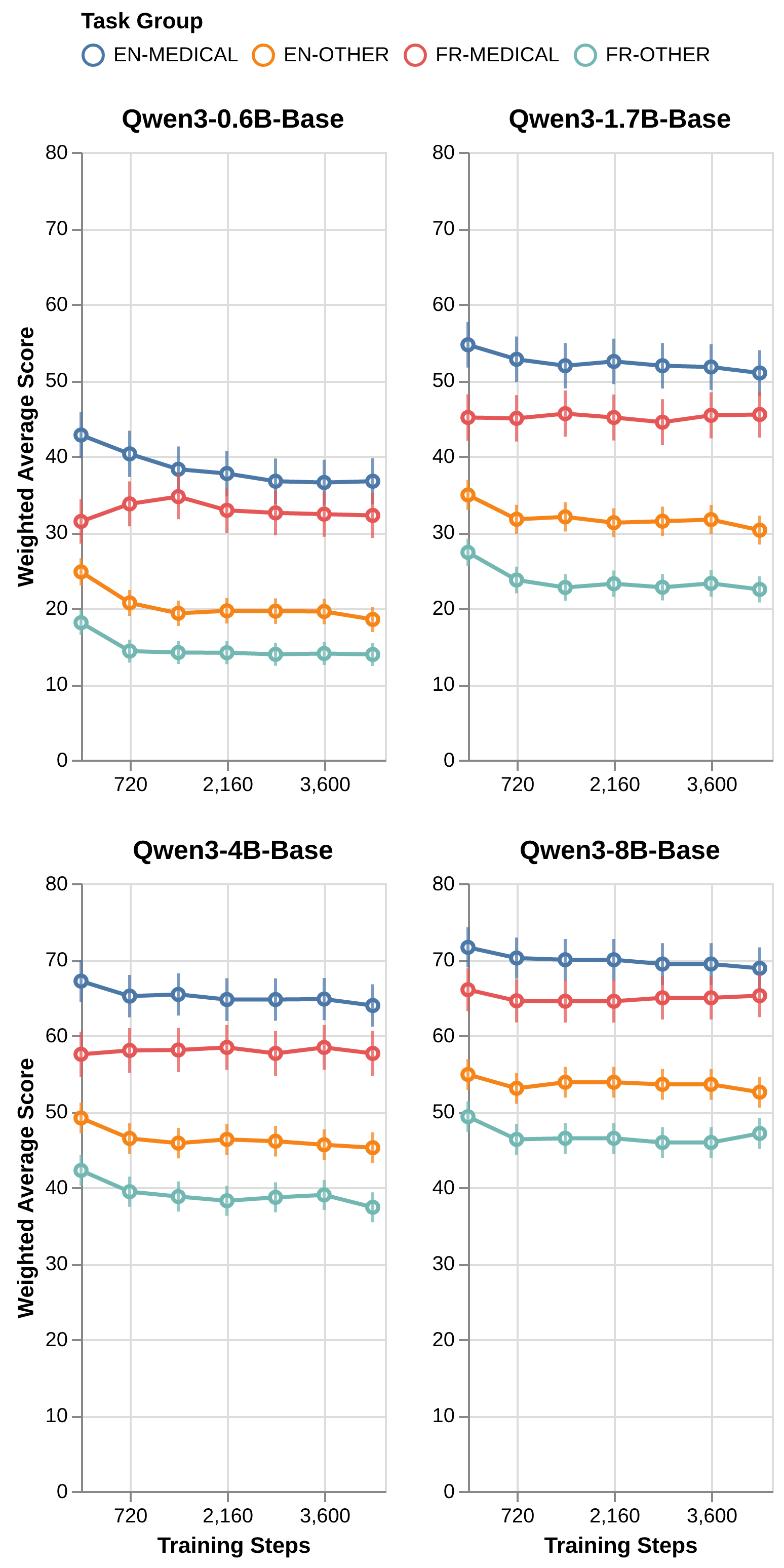}
        \caption{Progression of evaluation scores on the four task groups.}
        \label{fig:linechart}
    \end{center}
\end{figure}

Figure \ref{fig:linechart} displays the progression of the weighted average score over the PDAPT training process for each of the four members of the Qwen3 family considered.
As the MMLU-Pro-X datasets that make up the ``OTHER'' task groupings have more difficult questions in larger quantities (they were specifically designed to be more challenging than MMLU), and employ a more demanding evaluation metric (exact-match accuracy), the averages are significantly lower than for the medical-domain tasks.

We can see from these charts that the overall impact of PDAPT is minimal, with changes in the average becoming less pronounced as model size increases.
As would be expected, performance on the non-medical tasks decreases the more the models are exposed to the \corpusname corpus, although this is not necessarily accompanied by increases in medical-domain performance, and many of the averages trend back downward in the latter part of the training.
The only aspect of this that stands out as a potential avenue for improvement is the slight increase in the FR-MEDICAL average early in training for the smallest model, Qwen3-0.6B-Base.
Indeed, on further inspection, the 1440-step checkpoint gives us the greatest number of per-task improvements across all models.
It is thus these checkpoints for which the SLERP merging was carried out, and for which the task-by-task results are presented.

\paragraph{Baseline Evaluations}
For the task group that represents the target domain for the work in this project, FR-MEDICAL, we present a range of accuracy results for open-source LLMs in Appendix \ref{sec:apdx-benchmark}. 
These results provide a baseline reference for the performance metrics presented in Table \ref{tab:fr-medical}, by showing the metrics for both generalist and specialist models, with and without supervised training.
As they were beyond the range of parameter counts being considered for continual pre-training in this project, the 14B and 32B Qwen3 variants and GPT-oss-20B models are included for reference only.

\paragraph{Domain Adaptation Experiments} Side-by-side comparisons of the Qwen3 base models and their domain-adapted counterparts are presented for each task group as follows: Tables \ref{tab:fr-medical} and \ref{tab:en-medical} show results for the medical domain and Tables \ref{tab:fr-other} and \ref{tab:en-other} for the other specializations, for French and English respectively.
We highlight in \textbf{bold} results where the specialized models improved on the performance of the base model.
Green cells denote statistically significant increases (i.e. non-overlapping standard-error confidence intervals) and red cells statistically significant declines.

We thus make 56 comparisons per task domain: for FR-MEDICAL, there were 8 statistically significant changes, of which only 1 was negative.
For the other groups, the picture is somewhat more bleak - there was only one significant positive change (the model Qwen3-8B-Base+PDAPT+SLERP on the MMLU-Pro-X Business dataset in English).
However, it is worth noting that these declines apply to the PDAPT models only: once model merging is carried out, there are no longer any statistically significant decreases in performance for any of the base models across any of the task groups, while the performance on our actual target domain (FR-MEDICAL) remains elevated, particularly for the smaller models.

\begin{figure*}[ht]
    \begin{center}
        \includegraphics[width=2.075\columnwidth]{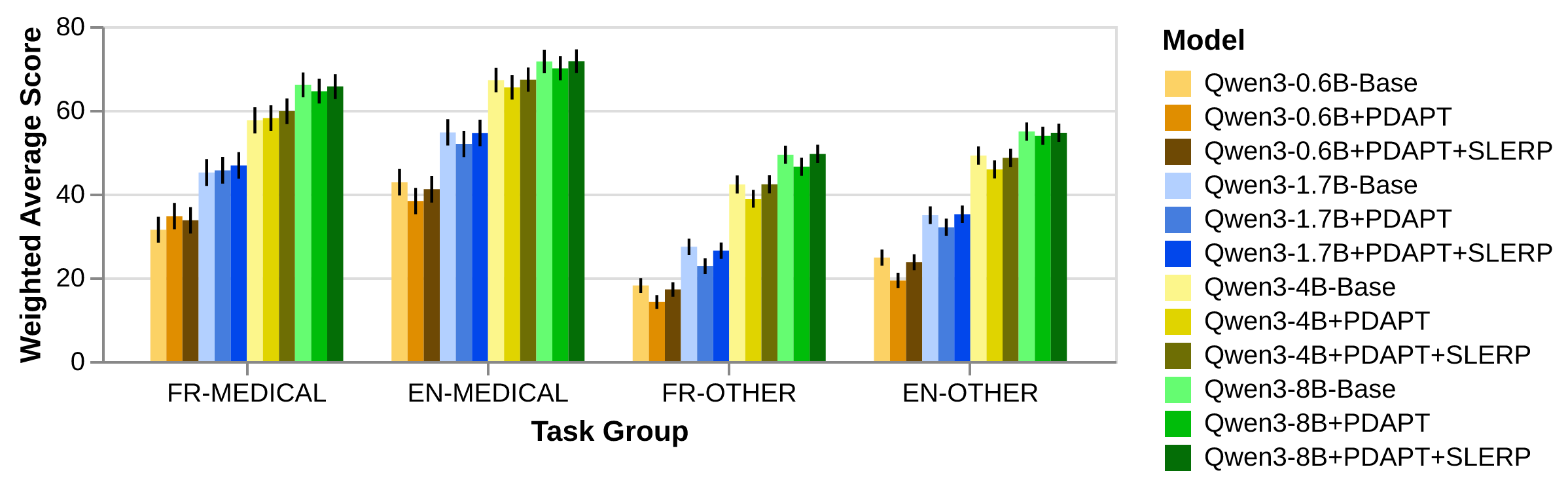}
        \caption{Comparison of group-level averages for base vs. specialized models, along with the SLERP merge between them. Black bars represent combined confidence intervals.}
        \label{fig:barchart}
    \end{center}
\end{figure*}

These results are summarized in the chart shown in Figure \ref{fig:barchart} - as observed in Figure \ref{fig:linechart}, we can see that there is little significant change in performance at the aggregate level - the per-task improvements in the FR-MEDICAL group appear to be cancelled out by concomitant losses in accuracy when averaged.
This suggests that DAPT is better approached in an even more specialized manner, at the level of medical subjects, and motivates further exploration into more granular experiments within the medical NLP domain.

\section{Conclusion}
\label{sec:conclusion}

This work introduces the \corpusname corpus, the first French biomedical corpus collection with full licensing compatibility for all downstream applications, addressing a gap in openly-available domain-specific resources.
Accompanying this corpus, we release the Qwen3-PDAPT collection, a series of decoder-only language models based on the Qwen3 pre-trained foundation models, which we hope will aid the research community in systematically assessing the capabilities of smaller language models in French biomedical contexts.
Through extensive experimentation and analysis encompassing multiple domains, we offer actionable insights and practical recommendations for researchers and practitioners working to adapt language models for healthcare applications in languages beyond English.

All datasets, models, training code and evaluation framework configurations used in this work, along with more extensive fully reproducible benchmarking experiments, will be made freely available online. 

As generalist language models continue to improve in quality and breadth, the marginal benefits of domain-adaptive pre-training are likely to diminish even further.
Nevertheless, given the more pronounced improvements we observed with smaller-scale decoder models, we advocate for continued investigation of DAPT in resource-constrained environments, where energy efficiency considerations are paramount; this is a particularly salient concern in healthcare settings that face stringent limitations on computational resources and restrictions on utilizing external model providers.
Furthermore, highly specialized pre-training targeting narrow biomedical subdomains may yield more substantial performance gains than broad domain adaptation.

Finally, our results demonstrate that merging domain-adapted models with their generalist base counterparts is not merely an optional enhancement but a fundamental requirement for maintaining balanced capabilities across both specialized and general language tasks.

Code for the experiments carried out in this work can be found at \githublink; the \corpusname corpus is available at \hdhlink.

\subsection{Limitations}

Our investigation focuses exclusively on causal language modeling as the pre-training objective and employs only the Qwen3 model family, without exploring supervised fine-tuning approaches that might complement domain adaptation.
The absence of publicly available technical specifications for Qwen3 constrained our ability to select optimal hyperparameters and training configurations for continued pre-training.
Moreover, our evaluation benchmark, while comprehensive, does not encompass the full breadth of medical subdomains and specialist topics necessary to thoroughly characterize the performance trade-offs inherent in domain-specific adaptation.

Although the SLERP model merging method was chosen for these experiments due to the fact that we are merging models with very similar loss trajectories, other merging techniques, notably DARE and/or TIES \citep{yadav_ties_2023}, represent promising alternatives to this method and have recently showed to leads to improved merged general and specialized models in specialized domains \citep{ueda2026merging}.

\subsection{Future Work}

As suggested in Section \ref{sec:results}, advancements of this work will involve the investigation of more targeted specialization by partitioning our pre-training corpora according to biomedical subtopics and applying selective continued pre-training to specific thematic subsets.
Exploring the effects of instruction fine-tuning on our domain-adapted models represents another crucial direction, as this technique may help reconcile specialized domain knowledge with general conversational capabilities.
Finally, we intend to expand our evaluation methodology beyond academic question-answering tasks to include practical document-processing and automation workflows encountered in real-world healthcare operations, thereby better assessing the utility of our models for applied clinical and administrative applications.

\section{Acknowledgements}

This work was carried out as part of the PARTAGES project, winner of the Bpifrance France 2030 call for proposals ``Digital Commons for Generative Artificial Intelligence''.
It was also partially supported by the French National Research Agency (ANR) through the MIAI ``AI \& Language'' chair (ANR-23-IACL-0006).
This work was performed using HPC resources from GENCI at IDRIS under allocation 2025-A0181016171 on the Jean Zay supercomputer.

\section{References}
\bibliographystyle{lrec2026-natbib}
\bibliography{biblio-main}

\appendix

\section{Appendix: Benchmarking Results}
\label{sec:apdx-benchmark}

\begin{table}[ht]
\setlength{\tabcolsep}{5pt}
\begin{tabularx}{0.45\textwidth}{l|r}
\textbf{Model} & \textbf{Average Ranking} \\ \hline
Qwen3-32B                & 1.57 \\
Qwen3-14B                & 2.43 \\
MedGemma-27B-it          & 2.71 \\
Qwen3-8B-Base            & 3.71 \\
Qwen3-8B                 & 4.86 \\
Qwen3-4B                 & 6.86 \\
Qwen3-4B-Base            & 7.43 \\
Apertus-8B-Instruct      & 11.14 \\
Apertus-8B               & 11.71 \\
Llama3.1-8B              & 12.57 \\
GPT-OSS-20B              & 13.00 \\
Apollo-7B                & 13.57 \\
Llama3.1-8B-Instruct     & 13.71 \\
EuroLLM-9B-Instruct      & 13.86 \\
MedGemma-4B-it           & 15.14 \\
Qwen3-1.7B-Base          & 15.43 \\
EuroLLM-9B               & 16.43 \\
Mistral-7B-Instruct-v0.3 & 18.00 \\
SmolLM-3B                & 18.29 \\
Mistral-7B-v0.3          & 18.43 \\
Llama3.2-1B              & 20.14 \\
Olmo3-7B                 & 21.43 \\
Qwen3-1.7B               & 21.71 \\
Gaperon-8B               & 22.57 \\
BioMistral-7B            & 24.14 \\
MedGemma-4B-pt           & 25.14 \\
Qwen3-0.6B-Base          & 26.71 \\
Qwen3-0.6B               & 27.00 \\
Llama3.2-1B-Instruct     & 27.57 \\
Gemma3-1B-it             & 29.00 \\
Gemma3-1B-pt             & 32.00 \\
Gaperon-1B               & 32.29 \\
Gemma3-270m-it           & 32.43 \\
Baguettotron             & 33.14 \\
Gemma3-270m-pt           & 33.86
\end{tabularx}
\caption{Average ranking of all 35 models across all seven French-language tasks.}
\label{tab:avg_rankings_apdx}
\end{table}


This section lays out the full results of the comparative benchmark used for model selection, i.e. the FR-MEDICAL task group.
Tables \ref{tab:mmlumed1_results_apdx}-\ref{tab:mmluproxmed_results_apdx} detail the evaluation metrics, alongside the corresponding EN-MEDICAL scores for reference. 
The top FR-MEDICAL result is highlighted in green and B text denotes all scores whose confidence intervals overlap with it.

In addition to the Qwen3 models presented in Section \ref{sec:training}, our evaluations include models from Google's Gemma3 \citep{gemmateam2025gemma3technicalreport} and MedGemma \citep{sellergren2025medgemmatechnicalreport} families, Meta's LLama3 family \citep{grattafiori2024llama3herdmodels}, and Mistral's 7B models \citep{jiang2023mistral7b}.
As mentioned in Section \ref{sec:training}, we include specialised biomedical models Apollo-7B \citep{zheng_efficiently_2024}, and BioMistral \citep{labrak_biomistral_2024}.
Alongside these models that are trained at private companies and research labs and are open-source only in the sense that their weights are freely available online, we include fully open models (base and instruction-tuned) EuroLLM-9B \citep{martins2025eurollm9btechnicalreport}, Apertus-8B \citep{apertus2025apertusdemocratizingopencompliant}, Gaperon-8B \citep{godey2025gaperonpepperedenglishfrenchgenerative}, and SmolLM-3B \citep{bakouch2025smollm3}.

In addition, we include results from GPT-OSS-20B \citep{openai2025gptoss120bgptoss20bmodel}, although on inspection of its outputs it appears that additional safety guardrail training inhibits the propensity of this particular model to give explicit answers to many medical questions.

Table \ref{tab:avg_rankings_apdx} shows the average rank of each model across the seven tasks.

The benchmark results underline the centrality of model size as a determining factor in performance, unsurprising for knowledge-base tasks such as these, with the top three models being the three largest in terms of parameter count.
However, it also underlines the aforementioned ability of the Qwen3 models to punch above their weight, with Qwen3-8B-Base coming within a statistically insignificant margin of these larger models in four out of seven tasks, and the Qwen3-4B models outperforming all of the 7-9B models apart from its own larger version.

Given resource and operational constraints in the PARTAGES project, the largest models considered for continual pre-training on \corpusname were in the 7-9B range. 

As we use an evaluation setup that directly accesses the log-likelihood distributions output by the decoders, we can see that instruction tuning is not necessarily helpful in this benchmark, although variations in the amount and type of supervised fine-tuning involved in building the instruction-tuned version from the base version also seem to play a role.
As noted previously, the Qwen3 models' ``-Base'' versions rank higher than their instruction-tuned counterparts (apart from the case of the 4B model) while the instruction-tuned Gemma3 and MedGemma instruction-tuned models (denoted by the ``-it'' suffix) all significantly improve on the performance of the corresponding base models (``-pt'' suffix).


\begin{table*}[ht]
\centering
\setlength{\tabcolsep}{5pt}
\begin{tabularx}{0.85\textwidth}{l|lcccc}
\textbf{Type} & \textbf{Model} & \multicolumn{4}{c}{\textbf{Accuracy (\%)$\uparrow$}} \\
& & \multicolumn{2}{c}{\textbf{Anatomy}} & \multicolumn{2}{c}{\textbf{Clinical Knowledge}} \\
& & \textbf{EN} & \textbf{FR} & \textbf{EN} & \textbf{FR} \\
\hline
\multirow{15}{*}{Base}
& Gemma3-270m-pt & 16.30$\pm$3.2 & 16.30$\pm$3.2 & 28.68$\pm$2.8 & 23.77$\pm$2.6 \\
& Baguettotron & 20.00$\pm$3.5 & 26.67$\pm$3.8 & 24.53$\pm$2.6 & 24.53$\pm$2.6 \\
& Qwen3-0.6B-Base & 47.41$\pm$4.3 & 29.63$\pm$3.9 & 56.98$\pm$3.0 & 49.81$\pm$3.1 \\
& Llama3.2-1B & 40.74$\pm$4.2 & 26.67$\pm$3.8 & 30.57$\pm$2.8 & 27.55$\pm$2.8 \\
& Gemma3-1B-pt & 32.59$\pm$4.0 & 28.89$\pm$3.9 & 21.89$\pm$2.5 & 23.77$\pm$2.6 \\
& Gaperon-1B & 17.04$\pm$3.2 & 19.26$\pm$3.4 & 26.79$\pm$2.7 & 32.08$\pm$2.9 \\
& Qwen3-1.7B-Base & 59.26$\pm$4.2 & 48.15$\pm$4.3 & 67.92$\pm$2.9 & 59.25$\pm$3.0 \\
& Qwen3-4B-Base & 68.15$\pm$4.0 & 54.81$\pm$4.3 & 80.38$\pm$2.4 & 70.94$\pm$2.8 \\
& Olmo3-7B & 57.78$\pm$4.3 & 44.44$\pm$4.3 & 69.43$\pm$2.8 & 55.85$\pm$3.1  \\
& Mistral-7B-v0.3 & 61.48$\pm$4.2 & 55.56$\pm$4.3 & 65.66$\pm$2.9 & 59.25$\pm$3.0 \\
& Qwen3-8B-Base & 74.07$\pm$3.8 & \textbf{62.22$\pm$4.2} & 80.00$\pm$2.5 & 74.72$\pm$2.7 \\
& Llama3.1-8B & 60.74$\pm$4.2 & 57.78$\pm$4.3 & 72.08$\pm$2.8 & 62.26$\pm$3.0 \\
& Apertus-8B & 60.00$\pm$4.2 & 54.81$\pm$4.3 & 73.21$\pm$2.7 & 66.42$\pm$2.9 \\
& Gaperon-8B & 54.81$\pm$4.3 & 40.00$\pm$4.2 & 55.47$\pm$3.1 & 54.72$\pm$3.1 \\
& EuroLLM-9B & 57.04$\pm$4.3 & 56.30$\pm$4.3 & 61.51$\pm$3.0 & 59.25$\pm$3.0 \\
\hline
\multirow{15}{*}{Instruct}
& Gemma3-270m-it & 26.67$\pm$3.8 & 22.96$\pm$3.6 & 25.28$\pm$2.7 & 28.30$\pm$2.8 \\
& Qwen3-0.6B & 48.15$\pm$4.3 & 45.19$\pm$4.3 & 50.19$\pm$3.1 & 44.53$\pm$3.1 \\
& Llama3.2-1B-Instruct & 51.85$\pm$4.3 & 44.44$\pm$4.3 & 47.17$\pm$3.1 & 38.87$\pm$3.0 \\
& Gemma3-1B-it & 41.48$\pm$4.3 & 34.07$\pm$4.1 & 43.77$\pm$3.1 & 43.77$\pm$3.1 \\
& Qwen3-1.7B & 57.04$\pm$4.3 & 40.74$\pm$4.2 & 63.02$\pm$3.0 & 55.09$\pm$3.1 \\
& SmolLM-3B & 54.81$\pm$4.3 & 45.93$\pm$4.3 & 67.55$\pm$2.9 & 60.75$\pm$3.0 \\
& Qwen3-4B & 63.70$\pm$4.2 & 57.78$\pm$4.3 & 76.23$\pm$2.6 & 68.30$\pm$2.9 \\
& Mistral-7B-Instruct-v0.3 & 64.44$\pm$4.1 & 44.44$\pm$4.3 & 68.30$\pm$2.9 & 60.38$\pm$3.0 \\
& Qwen3-8B & 73.33$\pm$3.8 & \textbf{62.96$\pm$4.2} & 77.36$\pm$2.6 & 74.72$\pm$2.7 \\
& Llama3.1-8B-Instruct & 51.85$\pm$4.3 & 44.44$\pm$4.3 & 47.17$\pm$3.1 & 38.87$\pm$3.0 \\
& Apertus-8B-Instruct & 59.26$\pm$4.2 & \textbf{60.74$\pm$4.2} & 70.94$\pm$2.8 & 61.89$\pm$3.0  \\
& EuroLLM-9B-Instruct & 59.26$\pm$4.2 & 57.04$\pm$4.3 & 60.38$\pm$3.0 & 59.62$\pm$3.0 \\
& Qwen3-14B & 79.26$\pm$3.5 & \textbf{67.41$\pm$4.0} & 81.51$\pm$2.4 & \textbf{78.49$\pm$2.5} \\
& GPT-OSS-20B & 48.89$\pm$4.3 & 60.00$\pm$4.2 & 66.42$\pm$2.9 & 62.26$\pm$3.0  \\
& Qwen3-32B & 80.00$\pm$3.5 & \textbf{67.41$\pm$4.0} & 86.42$\pm$2.1 & \cellcolor{green!25}\textbf{80.38$\pm$2.4}  \\
\hline
\multirow{5}{*}{BioMed}
& MedGemma-4B-pt & 48.89$\pm$4.3 & 42.96$\pm$4.3 & 52.83$\pm$3.1 & 48.68$\pm$3.1 \\
& MedGemma-4B-it & 54.81$\pm$4.3 & 52.59$\pm$4.3 & 61.51$\pm$3.0 & 63.02$\pm$3.0 \\
& Apollo-7B & 59.26$\pm$4.2 & 47.41$\pm$4.3 & 69.43$\pm$2.8 & 61.51$\pm$3.0 \\
& BioMistral-7B & 46.67$\pm$4.3 & 42.96$\pm$4.3 & 63.77$\pm$3.0 & 55.47$\pm$3.1 \\
& MedGemma-27B-it & 79.26$\pm$3.5 & \cellcolor{green!25}\textbf{68.89$\pm$4.0} & 81.89$\pm$2.4 & \textbf{76.98$\pm$2.6}
\end{tabularx}
\caption{Accuracy (mean) and standard error on MMLU medical benchmarks.}
\label{tab:mmlumed1_results_apdx}
\end{table*}

\begin{table*}[ht]
\centering
\setlength{\tabcolsep}{5pt}
\begin{tabularx}{0.85\textwidth}{l|lcccc}
\textbf{Type} & \textbf{Model} & \multicolumn{4}{c}{\textbf{Accuracy (\%)$\uparrow$}} \\
& & \multicolumn{2}{c}{\textbf{College Biology}} & \multicolumn{2}{c}{\textbf{College Medecine}} \\
& & \textbf{EN} & \textbf{FR} & \textbf{EN} & \textbf{FR} \\
\hline
\multirow{15}{*}{Base}
& Gemma3-270m-pt & 20.83$\pm$3.4 & 21.53$\pm$3.4 & 23.70$\pm$3.2 & 19.65$\pm$3.0 \\
& Baguettotron & 26.39$\pm$3.7 & 18.06$\pm$3.2 & 27.17$\pm$3.4 & 23.12$\pm$3.2 \\
& Qwen3-0.6B-Base & 59.72$\pm$4.1 & 39.58$\pm$4.1 & 52.60$\pm$3.8 & 40.46$\pm$3.7 \\
& Llama3.2-1B & 77.08$\pm$3.5 & 62.50$\pm$4.1 & 64.74$\pm$3.6 & 59.54$\pm$3.7 \\
& Gemma3-1B-pt & 25.69$\pm$3.7 & 27.08$\pm$3.7 & 21.39$\pm$3.1 & 21.97$\pm$3.2 \\
& Gaperon-1B & 22.92$\pm$3.5 & 20.14$\pm$3.4 & 20.81$\pm$3.1 & 21.39$\pm$3.1 \\
& Qwen3-1.7B-Base & 72.92$\pm$3.7 & 54.17$\pm$4.2 & 68.21$\pm$3.6 & 59.54$\pm$3.7 \\
& Qwen3-4B-Base & 84.72$\pm$3.0 & 75.00$\pm$3.6 & 73.99$\pm$3.3 &  68.21$\pm$3.6 \\
& Olmo3-7B & 76.39$\pm$3.6 & 53.47$\pm$4.2 & 70.52$\pm$3.5 & 50.87$\pm$3.8  \\
& Mistral-7B-v0.3 & 70.14$\pm$3.8 & 54.86$\pm$4.2 & 63.01$\pm$3.7 & 53.18$\pm$3.8 \\
& Qwen3-8B-Base & 88.89$\pm$2.6 & \textbf{87.50$\pm$2.8} & 78.03$\pm$3.2 & \textbf{75.72$\pm$3.3} \\
& Llama3.1-8B & 77.08$\pm$3.5 & 62.50$\pm$4.1 & 64.74$\pm$3.6 & 59.54$\pm$3.7 \\
& Apertus-8B & 72.92$\pm$3.7 & 68.75$\pm$3.9 & 61.85$\pm$3.7 & 57.80$\pm$3.8 \\
& Gaperon-8B & 62.50$\pm$4.0 & 52.78$\pm$4.2 & 47.40$\pm$3.8 & 45.09$\pm$3.8 \\
& EuroLLM-9B & 66.67$\pm$3.9 & 66.67$\pm$3.9 & 54.34$\pm$3.8 & 52.02$\pm$3.8 \\
\hline
\multirow{15}{*}{Instruct}
& Gemma3-270m-it & 34.72$\pm$4.0 & 22.92$\pm$3.5 & 21.39$\pm$3.1 & 21.39$\pm$3.1 \\
& Qwen3-0.6B & 56.25$\pm$4.1 & 34.03$\pm$4.0 & 47.40$\pm$3.8 & 38.15$\pm$3.7 \\
& Llama3.2-1B-Instruct & 46.53$\pm$4.2 & 34.03$\pm$4.0 & 36.99$\pm$3.7 &  28.32$\pm$3.4 \\
& Gemma3-1B-it & 34.72$\pm$4.0 & 36.11$\pm$4.0 & 39.31$\pm$3.7 & 38.73$\pm$3.7 \\
& Qwen3-1.7B & 67.36$\pm$3.9 & 49.31$\pm$4.2 & 61.85$\pm$3.7 & 56.07$\pm$3.8 \\
& SmolLM-3B & 72.92$\pm$3.7 & 59.72$\pm$4.1 & 64.16$\pm$3.7 & 55.49$\pm$3.8 \\
& Qwen3-4B & 84.72$\pm$3.0 & 73.61$\pm$3.7 & 70.52$\pm$3.5 & \textbf{73.99$\pm$3.3} \\
& Mistral-7B-Instruct-v0.3 & 72.22$\pm$3.7 & 59.03$\pm$4.1 & 60.12$\pm$3.7 & 54.34$\pm$3.8 \\
& Qwen3-8B & 88.19$\pm$2.7 & 85.42$\pm$3.0 & 79.19$\pm$3.1 & \textbf{73.41$\pm$3.4} \\
& Llama3.1-8B-Instruct & 81.94$\pm$3.2 & 67.36$\pm$3.9 & 65.32$\pm$3.6 & 63.01$\pm$3.7 \\
& Apertus-8B-Instruct & 75.00$\pm$3.6 & 67.36$\pm$3.9 & 61.27$\pm$3.7 & 59.54$\pm$3.7 \\
& EuroLLM-9B-Instruct & 71.53$\pm$3.8 & 71.53$\pm$3.8 & 53.76$\pm$3.8 & 54.34$\pm$3.8 \\
& Qwen3-14B & 92.36$\pm$2.2 & \cellcolor{green!25}\textbf{89.58$\pm$2.6} & 80.92$\pm$3.0 & \cellcolor{green!25}\textbf{78.03$\pm$3.2} \\
& GPT-OSS-20B & 70.83$\pm$3.8 & 69.44$\pm$3.9 & 54.34$\pm$3.8 & 56.07$\pm$3.8  \\
& Qwen3-32B & 90.28$\pm$2.5 & 90.97$\pm$2.4 & 82.08$\pm$2.9 & \textbf{77.46$\pm$3.2}  \\
\hline
\multirow{5}{*}{BioMed}
& MedGemma-4B-pt & 59.03$\pm$4.1 & 48.61$\pm$4.2 & 49.71$\pm$3.8 & 42.77$\pm$3.8 \\
& MedGemma-4B-it & 68.75$\pm$3.9 & 54.86$\pm$4.2 & 55.49$\pm$3.8 & 56.65$\pm$3.8 \\
& Apollo-7B & 77.78$\pm$3.5 & 62.50$\pm$4.1 & 59.54$\pm$3.7 & 58.38$\pm$3.8 \\
& BioMistral-7B & 59.03$\pm$4.1 & 47.22$\pm$4.2 & 53.76$\pm$3.8 & 50.29$\pm$3.8 \\
& MedGemma-27B-it & 85.42$\pm$3.0 & \textbf{86.81$\pm$2.8} & 72.83$\pm$3.4 & \textbf{73.99$\pm$3.3}
\end{tabularx}
\caption{Accuracy (mean) and standard error on MMLU medical benchmarks.}
\label{tab:mmlumed2_results_apdx}
\end{table*}

\begin{table*}[ht]
\centering
\setlength{\tabcolsep}{5pt}
\begin{tabularx}{0.85\textwidth}{l|lcccc}
\textbf{Type} & \textbf{Model} & \multicolumn{4}{c}{\textbf{Accuracy (\%)$\uparrow$}} \\
& & \multicolumn{2}{c}{\textbf{Medical Genetics}} & \multicolumn{2}{c}{\textbf{Professional Medecine}} \\
& & \textbf{EN} & \textbf{FR} & \textbf{EN} & \textbf{FR} \\
\hline
\multirow{15}{*}{Base}
& Gemma3-270m-pt & 25.00$\pm$4.4 & 23.00$\pm$4.2 & 43.38$\pm$3.0 & 23.16$\pm$2.6 \\
& Baguettotron & 30.00$\pm$4.6 & 24.00$\pm$4.3 & 44.85$\pm$3.0 & 20.96$\pm$2.5 \\
& Qwen3-0.6B-Base & 62.00$\pm$4.9 & 47.00$\pm$5.0 & 55.51$\pm$3.0 & 37.13$\pm$2.9 \\
& Llama3.2-1B & 80.00$\pm$4.0 & 67.00$\pm$4.7 & 69.85$\pm$2.8 &  55.51$\pm$3.0 \\
& Gemma3-1B-pt & 27.00$\pm$4.5 & 26.00$\pm$4.4 & 30.51$\pm$2.8 & 24.63$\pm$2.6 \\
& Gaperon-1B & 30.00$\pm$4.6 & 27.00$\pm$4.5 & 44.85$\pm$3.0 &  30.88$\pm$2.8 \\
& Qwen3-1.7B-Base & 73.00$\pm$4.5 & 66.00$\pm$4.8 & 64.71$\pm$2.9 & 56.25$\pm$3.0 \\
& Qwen3-4B-Base & 81.00$\pm$3.9 & 74.00$\pm$4.4 & 78.31$\pm$2.5 & 69.49$\pm$2.8 \\
& Olmo3-7B & 76.00$\pm$4.3 & 63.00$\pm$4.9 & 63.60$\pm$2.9 & 48.90$\pm$3.0  \\
& Mistral-7B-v0.3 & 73.00$\pm$4.5 & 59.00$\pm$4.9 & 65.07$\pm$2.9 &  54.41$\pm$3.0 \\
& Qwen3-8B-Base & 86.00$\pm$3.5 & \textbf{80.00$\pm$4.0} & 83.46$\pm$2.3 & 76.47$\pm$2.6 \\
& Llama3.1-8B & 80.00$\pm$4.0 & 67.00$\pm$4.7 & 69.85$\pm$2.8 &  55.51$\pm$3.0 \\
& Apertus-8B & 67.00$\pm$4.7 & 64.00$\pm$4.8 & 60.29$\pm$3.0 & 59.56$\pm$3.0 \\
& Gaperon-8B & 62.50$\pm$4.0 & 65.00$\pm$4.8 & 48.53$\pm$3.0 & 42.28$\pm$3.0 \\
& EuroLLM-9B & 69.00$\pm$4.7 & 62.00$\pm$4.9 & 55.51$\pm$3.0 & 55.88$\pm$3.0 \\
\hline
\multirow{15}{*}{Instruct}
& Gemma3-270m-it & 27.00$\pm$4.5 & 28.00$\pm$4.5 & 36.76$\pm$2.9 & 18.75$\pm$2.4 \\
& Qwen3-0.6B & 54.00$\pm$5.0 & 38.00$\pm$4.9 & 41.18$\pm$3.0 & 33.46$\pm$2.9 \\
& Llama3.2-1B-Instruct & 57.00$\pm$5.0 & 46.00$\pm$5.0 & 51.84$\pm$3.0 &  26.47$\pm$2.7 \\
& Gemma3-1B-it & 41.00$\pm$4.9 & 44.00$\pm$5.0 & 27.21$\pm$2.7 & 20.96$\pm$2.5 \\
& Qwen3-1.7B & 75.00$\pm$4.4 & 62.00$\pm$4.9 & 58.09$\pm$3.0 & 48.16$\pm$3.0 \\
& SmolLM-3B & 67.00$\pm$4.7 & 58.00$\pm$5.0 & 55.88$\pm$3.0 & 49.26$\pm$3.0 \\
& Qwen3-4B & 80.00$\pm$4.0 & 69.00$\pm$4.6 & 76.10$\pm$2.6 & 68.38$\pm$2.8 \\
& Mistral-7B-Instruct-v0.3 & 72.22$\pm$3.7 & 60.00$\pm$4.9 & 62.50$\pm$2.9 & 52.94$\pm$3.0 \\
& Qwen3-8B & 86.00$\pm$3.5 & \textbf{80.00$\pm$4.0} & 82.35$\pm$2.3 &  72.79$\pm$2.7 \\
& Llama3.1-8B-Instruct & 84.00$\pm$3.7 & 68.00$\pm$4.7 & 76.47$\pm$2.6 &  61.40$\pm$3.0 \\
& Apertus-8B-Instruct & 65.00$\pm$4.8 & 62.00$\pm$4.9 & 62.87$\pm$2.9 & 60.66$\pm$3.0 \\
& EuroLLM-9B-Instruct & 63.00$\pm$4.9 & 64.00$\pm$4.8 & 59.56$\pm$3.0 & 56.25$\pm$3.0 \\
& Qwen3-14B & 89.00$\pm$3.1 & \textbf{80.00$\pm$4.0} & 83.46$\pm$2.3 & \textbf{78.68$\pm$2.5}  \\
& GPT-OSS-20B & 64.00$\pm$4.8 & 64.00$\pm$4.8 & 59.19$\pm$3.0 & 52.94$\pm$3.0 \\
& Qwen3-32B & 94.00$\pm$2.4 & \textbf{85.00$\pm$3.6} & 85.29$\pm$2.2 & \cellcolor{green!25}\textbf{84.19$\pm$2.2}  \\
\hline
\multirow{5}{*}{BioMed}
& MedGemma-4B-pt & 55.00$\pm$5.0 & 48.00$\pm$5.0 & 38.24$\pm$3.0 & 35.29$\pm$2.9 \\
& MedGemma-4B-it & 67.00$\pm$4.7 & 65.00$\pm$4.8 & 62.13$\pm$2.9 & 51.47$\pm$3.0 \\
& Apollo-7B & 77.00$\pm$4.2 & 67.00$\pm$4.7 & 68.01$\pm$2.8 &  59.93$\pm$3.0 \\
& BioMistral-7B & 65.00$\pm$4.8 & 48.00$\pm$5.0 & 55.15$\pm$3.0 &  46.32$\pm$3.0 \\
& MedGemma-27B-it & 87.00$\pm$3.4 & \cellcolor{green!25}\textbf{86.00$\pm$3.5} & 83.09$\pm$2.3 & \textbf{78.68$\pm$2.5}
\end{tabularx}
\caption{Accuracy (mean) and standard error on MMLU medical benchmarks.}
\label{tab:mmlumed3_results_apdx}
\end{table*}

\begin{table*}[ht]
\centering
\setlength{\tabcolsep}{5pt}
\begin{tabularx}{0.6\textwidth}{l|lcc}
\textbf{Type} & \textbf{Model} & \multicolumn{2}{c}{\textbf{Accuracy (\%)$\uparrow$}} \\
& & \textbf{EN} & \textbf{FR} \\
\hline
\multirow{15}{*}{Base}
& Gemma3-270m-pt & 9.02$\pm$1.1 & 9.46$\pm$1.1 \\
& Baguettotron & 10.92$\pm$1.2 & 6.26$\pm$0.9 \\
& Qwen3-0.6B-Base & 22.42$\pm$1.6 & 15.87$\pm$1.4 \\
& Llama3.2-1B & 14.85$\pm$1.4 & 10.77$\pm$1.2 \\
& Gemma3-1B-pt & 10.92$\pm$1.2 & 9.75$\pm$1.1 \\
& Gaperon-1B & 12.81$\pm$1.3 & 7.57$\pm$1.0 \\
& Qwen3-1.7B-Base & 34.64$\pm$1.8 & 25.91$\pm$1.7 \\
& Qwen3-4B-Base & 49.93$\pm$1.9 & 39.30$\pm$1.9 \\
& Olmo3-7B & 34.93$\pm$1.8 & 18.20$\pm$1.5 \\
& Mistral-7B-v0.3 & 35.52$\pm$1.8 & 27.22$\pm$1.7 \\
& Qwen3-8B-Base & 55.75$\pm$1.9 & 50.22$\pm$1.9 \\
& Llama3.1-8B & 43.38$\pm$1.9 & 28.53$\pm$1.7 \\
& Apertus-8B & 34.64$\pm$1.8 & 36.10$\pm$1.8 \\
& Gaperon-8B & 26.64$\pm$1.7 & 20.52$\pm$1.5 \\
& EuroLLM-9B & 33.92$\pm$1.8 & 29.11$\pm$1.7 \\
\hline
\multirow{15}{*}{Instruct}
& Gemma3-270m-it & 9.90$\pm$1.1 & 10.63$\pm$1.2 \\
& Qwen3-0.6B & 22.27$\pm$1.6 & 14.41$\pm$1.3 \\
& Qwen3-1.7B & 35.52$\pm$1.8 & 25.91$\pm$1.7 \\
& Llama3.2-1B-Instruct & 22.13$\pm$1.6 & 16.45$\pm$1.4 \\
& Gemma3-1B-it & 16.16$\pm$1.4 & 13.68$\pm$1.3 \\
& SmolLM-3B & 36.97$\pm$1.8 & 29.40$\pm$1.7 \\
& Qwen3-4B & 52.11$\pm$1.9 & 41.63$\pm$1.9 \\
& Mistral-7B-Instruct-v0.3 & 40.90$\pm$1.9 & 34.21$\pm$1.8 \\
& Qwen3-8B &  57.50$\pm$1.9 & 48.18$\pm$1.9 \\
& Llama3.1-8B-Instruct & 49.64$\pm$1.9 & 36.39$\pm$1.8 \\
& Apertus-8B-Instruct & 41.48$\pm$1.9 & 41.92$\pm$1.9 \\
& EuroLLM-9B-Instruct & 36.97$\pm$1.8 & 30.86$\pm$1.8 \\
& Qwen3-14B & 63.03$\pm$1.8 & 56.48$\pm$1.9 \\
& GPT-OSS-20B & 37.26$\pm$1.9 & 35.81$\pm$1.8 \\
& Qwen3-32B & 68.85$\pm$1.8 & \cellcolor{green!25}\textbf{66.08$\pm$1.8} \\
\hline
\multirow{5}{*}{BioMed}
& MedGemma-4B-pt & 20.67$\pm$1.5 & 18.49$\pm$1.5 \\
& MedGemma-4B-it & 41.48$\pm$1.9 & 31.30$\pm$1.8 \\
& Apollo-7B & 43.67$\pm$1.9 & 30.71$\pm$1.8 \\
& BioMistral-7B & 32.17$\pm$1.8 & 20.23$\pm$1.5 \\
& MedGemma-27B-it & 65.36$\pm$1.8 & 61.86$\pm$1.9
\end{tabularx}
\caption{Accuracy (mean) and standard error on the MMLU-Pro-X task \textit{Health}.}
\label{tab:mmluproxmed_results_apdx}
\end{table*}

\end{document}